\documentclass[sigconf]{acmart}
\usepackage{amsthm}
\usepackage{amsmath}
\usepackage{amsmath}

\usepackage{algorithm,algcompatible}
\AtBeginDocument{%
  \providecommand\BibTeX{{%
    \normalfont B\kern-0.5em{\scshape i\kern-0.25em b}\kern-0.8em\TeX}}}
\setcopyright{acmcopyright}
\copyrightyear{2018}
\acmYear{2018}
\acmDOI{10.1145/1122445.1122456}

\acmConference[Atlanta '22]{Atlanta '22: 31st ACM International Conference on Information and Knowledge Management}{October 17--22, 2022}{Atlanta, USA}
\acmPrice{15.00}
\acmISBN{978-1-4503-XXXX-X/18/06}

\begin{document}

\title{Real-time Anomaly Detection for Multivariate Data Streams}

\author{Kenneth Odoh}
\affiliation{\institution{Microsoft Corporation}}
\email{kenneth.odoh@gmail.com}

\begin{abstract}

We present a real-time multivariate anomaly detection algorithm for data streams based on the Probabilistic Exponentially Weighted Moving Average (PEWMA). Our formulation is resilient to (abrupt transient, abrupt distributional, and gradual distributional) shifts in the data. The novel anomaly detection routines utilize an incremental online algorithm to handle streams. Furthermore, our proposed anomaly detection algorithm works in an unsupervised manner eliminating the need for labeled examples. Our algorithm performs well and is resilient in the face of concept drifts.
\end{abstract}




\maketitle

\section{Introduction}

Anomaly detection is the task of classifying patterns that depict abnormal behavior. Outliers can arise due to (human/equipment) errors, faulty systems, and others. Anomaly detection is well-suited for unbalanced data, where the ideal scenario is to predict the behavior of the minority class. There are numerous applications in detecting default on loans, fraud detection, and network intrusion detections. An anomaly detection algorithm can work in Unsupervised, Supervised, or hybrid mode.

There are different types of anomalies described as follows.
\begin{itemize}
\item Point Anomaly: the algorithm decide a single instance as an anomaly concerning the entire data set. 
\item Contextual Anomaly: a data instance can be anomalous based on the context (attributes and position in the stream) and proximity of the chosen anomaly. This anomaly type is ideal for multivariate data, e.g. in the snapshot reading of a machine, an attribute of a single data point may seem abnormal but can be normal behavior based on consideration of the entire data.
\item Collective Anomaly: the algorithm decide a set of data points that are anomalies as a group, but individually these data points exhibit normal behaviors.
\end{itemize}

Anomaly detection algorithms can operate in the following setting:
\begin{itemize}
\item Static: These algorithms work in static datasets. Every item is loaded into memory at one time to perform computation.
\item Online: These algorithms work in real-time data streams. Items are incrementally loaded into memory and processed in chunks.
\item Static + Online: The model may operate in two stages as initial parameters get estimated in the static setting. The parameters are incrementally updated as more data arrives. Our work is of this type.
\end{itemize}

PEWMA was introduced in the work~\cite{Carter2012} for online anomaly detection on univariate time series. Drawing inspiration from their work, we have provided extensions to support real-time anomaly detection of a multivariate data stream. 

\footnote{Source code: \url{https://github.com/kenluck2001/anomalyMulti}, Blog: \url{https://kenluck2001.github.io/blog_post/real-time_anomaly_detection_for_multivariate_data_stream.html}}

\section{Background}
\begin{figure}[H] 
\centering
\includegraphics[scale=0.45]{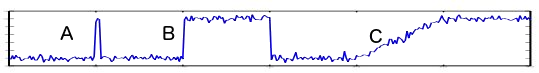}
\caption{Different types of signal changes: abrupt transient, abrupt distributional shift, and gradual distributional shift~\cite{Carter2012} (from left to right). }
\label{anomalytype}
\end{figure}

An anomaly detection algorithm can identify outliers in several signal changes in time-dependent data. 
Signal changes are common in time-dependent data. They include abrupt transient shift, abrupt distributional shift, and gradual distributional shift~\cite{Carter2012} labeled as "A", "B", and "C"  in Figure~\ref{anomalytype} respectively.

Online algorithms are useful for real-time applications, as they operate incrementally on data streams. These algorithms incrementally receive input and decide based on an updated parameter that conveys the current state of the data stream. This philosophy contrasts with offline algorithms that assume the entire data is available in memory. The issue with an offline algorithm is that the data must fit in memory. The online algorithm should be both time and space-efficient.

Anomaly detection algorithm can work in modes such as diagnosis and accommodation~\cite{hodge2004}. Firstly, the diagnosis method finds the outlier in the data and removes it from the data sample to avoid skewing the distribution. This method is applicable when the distribution of expected behaviors is known. The outliers get excluded when the estimation of the parameters of the distribution~\cite{hodge2004}. Secondly, the accommodation method finds the outliers in the data and includes estimating the parameters of the statistical model. The accommodation method is suitable for data streams that account for the effect of concept drift~\cite{Ditzler2013}.

Muth~\cite{muth60} laid the foundation for exponential smoothing by showing that it provides optimal estimates using random walk with some noise. Further works were done to provide a statistical framework for exponential smoothing leading to the creation of linear models~\cite{boxjen76, roberts82}. Unfortunately, this does apply to nonlinear models. Yule postulated a formulation of time series as a realization of a stochastic process~\cite{yule27}. Exponential Weighted Moving Average (EWMA) is ideal for keeping a set of running moments in the data stream. EWMA is a smoothing technique that adds a forgetting parameter to modify the influence of a recent item in the data stream as shown in Equation~\ref{emwa_equation}. This smoothing causes volatility in abrupt transient changes and is unfit for distribution shifts.

\begin{equation}
\label{emwa_equation}
\mu_{t+1} = \alpha\mu_{t-1} + (1 - \alpha) X_{t}
\end{equation}

EWMA limitations motivated the discovery of Probabilistic Exponentially Weighted Moving Average (PEWMA). PEWMA~\cite{Carter2012} improves on EMWA by adding parameter which includes the probability of the data in the model as shown in Equation~\ref{pewma_equation}. PEWMA works for every shift including abrupt transient shift, abrupt distributional shift, and gradual distributional shift respectively. PEWMA and EWMA make use of a damped window~\cite{Salehi2018}.

\begin{equation}
\label{pewma_equation}
\mu_{t+1} = \alpha(1 - \beta P_{t}) \mu_{t-1} + (1 - \alpha(1 - \beta P_{t}) ) X_{t}
\end{equation}

\section{Method}
Our formulation provided an implementation of the online covariance matrix in Subsection~\ref{online-covariance-matrix}, alongside an online inverse covariance matrix based on Sherman$-$Morrison formula in Subsection~\ref{online-inverse-covariance-matrix}, and PEWMA in Subsection~\ref{prob-exp-weighted-avg}. 

Our implementation of the online covariance matrix builds on work~\cite{Igel2006}. We simplify the algorithm by ignoring the details of evolutionary computation in the paper. Our work adapted evolution as described in the paper~\cite{Igel2006} as a transition from one generation to the next; is equivalent to moving from one state to another state. This is analogous to how online algorithms work with dynamic changes as new data enters the stream. Cholesky decomposition is used extensively in the algorithms. 

\subsection{Probabilistic Exponentially Weighted Moving Average (PEWMA)}
\label{prob-exp-weighted-avg}

PEWMA~\cite{Carter2012} algorithm works in the accommodation mode. The routine shown in Algorithm [1] allows for concept drift~\cite{Ditzler2013}, which occurs in data streams by updating the set of parameters that convey the state of the stream.

\begin{algorithm}
\caption{Probabilistic Exponential Weighted Moving Average~\cite{Carter2012}}
\begin{algorithmic} 

\REQUIRE $X_{t},  \hat{X_{t}}, \hat{\alpha_{t}}, T, t$    
\ENSURE $P_{t}, \hat{X_{t+1}}, \hat{\alpha_{t+1}} $  

\begin{verbatim}
//incremental Z score
\end{verbatim}
\STATE $Z_{t} \leftarrow \frac{ X_{t} - \hat{X_{t}} }  {\hat{\alpha_{t}}}$

\begin{verbatim}
//probability density function
\end{verbatim}

\STATE $P_{t} \leftarrow     \frac{ Z_{t} }  {  \sqrt{2\pi} }   e^{ \frac{ Z_{t} }  { 2 } }$

\IF{$t < T$}

\begin{verbatim}
//increment standard deviation (training phase)
\end{verbatim}
\STATE $\alpha_{t} \leftarrow 1 - 1/t$ 
\ELSE

\begin{verbatim}
//increment standard deviation
\end{verbatim}
\STATE $\alpha_{t} \leftarrow (1 - \beta P_{t} ) \alpha$ 
\ENDIF

\begin{verbatim}
//moving average
\end{verbatim}
\STATE $s_{1} \leftarrow \alpha_{t} s_{1} + (1 - \alpha_{t} )X_{t}$
\STATE $s_{2}  \leftarrow \alpha_{t} s_{2} + (1 - \alpha_{t} )X_{t}^2$ 

\begin{verbatim}
//incremental mean
\end{verbatim}
\STATE $\hat{X_{t+1}} \leftarrow s_{1}$ 

\begin{verbatim}
//incremental standard deviation
\end{verbatim}
\STATE $\hat{\alpha_{t+1}} \leftarrow \sqrt{ s_{2} - s_{1}^2  }$  

\label{anomalgo }
\end{algorithmic}
\end{algorithm}

The parameters of the anomaly detection algorithm consist of $X_{t}$ the current data, $\mu_{t}$ the mean of the data, $\hat{X_{t}}$ is the mean of the data, $\hat{\alpha_{t}}$ the current standard deviation, $P_{t}$ the probability density function, $\hat{X_{t+1}}$ the mean of the next data (incremental aggregate), $\hat{\alpha_{t+1}}$ the next standard deviation (incremental aggregates), $T$ the data size, and $t$ a point in $T$. Initialize the process by setting the initial data for training the model $s_{1} = X_{1}$ and $s_{2} = X_{1}^{2}$.

Our work made use of the following parameters $\alpha = 0.98, \beta = 0.98$, and $\tau = 0.0044$. These thresholds are chosen based on the criteria that outliers are $\geq 3$ times the standard deviation in normally distributed data.

\subsection{Online Covariance matrix}
\label{online-covariance-matrix}
\begin{enumerate}
\item Estimate covariance matrix for initial data, $X \in R^{n \times m}$.
Initial covariance matrix, $C$ where $C \in R^{n \times m}$, $n$ is the number of samples, $m$ is the number of dimensions as shown in Equation~\ref{cov_equation}.
    
\begin{equation}
\label{cov_equation}
C = X * {X}^T
\end{equation}
    
\item Perform Cholesky factorization on the initial covariance matrix (positive-definite), $C$ as shown in Equation~\ref{cov_chol_equation}.
\begin{equation}
\label{cov_chol_equation}
C_t = A_t * {A_t}^T
\end{equation}

\item The updated covariance in the presence of new data is equivalent to the weighted average of the past covariance without the updated data and covariance of the transformed input as shown in Equation~\ref{xxincr_cov_chol_equation}.

\begin{equation}
\label{xxincr_cov_chol_equation}
C_{t+1} = \alpha * C_t + \beta * v_t * {v_t}^T
\end{equation}

Where $v_t = A_t * z_t$ and $z_t \in R^m$ is understood in our implementation is the current data. $\alpha$ and $\beta$ are positive scalar values.

\item Increment the Cholesky factor of the covariance matrix as shown in Equation~\ref{incr_cov_chol_coeff}.

\begin{equation}
\label{incr_cov_chol_coeff}
A_{t+1} = \sqrt{\alpha} * A_t + \frac{\sqrt{\alpha}}{\Big\|z_t \Big\|^2} * \left( \sqrt{1 + \frac{\beta * \Big\|z_t \Big\|^2}{\alpha}} - 1 \right) * v_t * z_t
\end{equation}

\item There are difficulties with setting the values of $\alpha$ and $\beta$ respectively. $\alpha + \beta = 1$ as an explicit form of exponential moving average coefficients. The author chose to set the values of $\alpha$, $\beta$ using the statistics of the data stream as shown in Equation~\ref{incr_cov_chol_coeff2}.

The parameters are set as $\alpha = {C_{a}}^2$, $\beta = 1 - {C_{a}}^2$ and $n$ is the size of the original data in the static settings.
Where ${C_{a}} = \sqrt{1 - C_{cov}}$ and $C_{cov} = \frac{2}{{n^2}+6}$.

\begin{equation}
\label{incr_cov_chol_coeff2}
A_{t+1} = {C_{a}} * A_t + \frac{{C_{a}}}{\Big\|z_t \Big\|^2} * \left( \sqrt{1 + \frac{(1 - {C_{a}}^2) * \Big\|z_t \Big\|^2}{{C_{a}}^2}} - 1 \right) * v_t * z_t
\end{equation}

\end{enumerate}

\subsection{Online Inverse Covariance matrix}
\label{online-inverse-covariance-matrix}
\begin{enumerate}
\item Estimate covariance matrix for initial data, $X \in R^{n \times m}$.
Initial covariance matrix, $C$ where $C \in R^{n \times m}$, $n$ is the number of samples, $m$ is the number of dimensions as shown in Equation~\ref{inverse_cov_equation}.
    
\begin{equation}
\label{inverse_cov_equation}
C = X * {X}^T
\end{equation}

Inverse the covariance matrix, $C^{-1}$ as shown in Equation~\ref{inverse_cov_equation_real}.

\begin{equation}
\label{inverse_cov_equation_real}
C^{-1} = \left( X * {X}^T \right)^{-1}
\end{equation}
    
\item Perform Cholesky factorization on initial covariance matrix, $C$ as shown in Equation~\ref{inverse_cov_chol_equation_real}.
\begin{equation}
\label{inverse_cov_chol_equation_real}
C_t = A_t * {A_t}^T
\end{equation}

\item Increment the Cholesky factor of the covariance matrix

\begin{equation}
\label{incr_cov_chol_equation1}
C_{t+1}^{-1} = (\alpha * C_t + \beta * v_t * {v_t}^T)^{-1}
\end{equation}

\begin{equation}
\label{incr_cov_chol_equation2}
C_{t+1}^{-1} = \alpha^{-1} * (C_t + \frac{\beta * v_t * {v_t}^T}{\alpha})^{-1}
\end{equation}

Let us fix, $\hat{v_t} = \frac{\beta * v_t}{\alpha}$. The resulting simplification using Sherman$-$Morrison Formula reduces the expression to

\begin{equation}
\label{incr_cov_chol_equation3}
C_{t+1}^{-1} = \frac{1}{\alpha} * \left({{C_t}^{-1}} - \frac{{{C_t}^{-1}} * \hat{v_t} * {v_t}^T * {{C_t}^{-1}}}{1 + (\hat{v_t} * {{C_t}^{-1}} * {v_t}^T)} \right)
\end{equation}
\end{enumerate}

\subsection{Online Multivariate Anomaly Detection}

The probability density function utilizes ideas from hypothesis testing for Deciding on a threshold to set the confidence level for determining the acceptance and rejection regions of the Gaussian distribution curve.

\begin{enumerate}
\item Use the covariance matrix, $C_{t+1}$ and inverse covariance matrix, ${C_{t+1}}^{-1}$.

\item We increment the mean vector, $\mu$ as new data arrives. It is possible to simplify the Covariance matrix, $C$, which will capture a number of the dynamics of the system. Let $n$ represent the current count of data before new data has arrived. Also, $\hat{x}$: is the new data, $\mu_{t+1}$: moving average as shown in Equation~\ref{incr_mean}.
\begin{equation}
\label{incr_mean}
\mu_{t+1} = \frac{(n * \mu_t) + \hat{x}}{n+1}
\end{equation}

\item Set a threshold to determine the acceptance and rejection regions. Items in the acceptance region are considered to be normal behavior as shown in Equation~\ref{pdf_equation}.
\begin{equation}
\label{pdf_equation}
p(x)=\frac{1}{\sqrt{(2\pi)^m|C|}} \exp\left(-\frac{1}{2}(x-\mu)^T{C}^{-1}(x-\mu) \right)
\end{equation}
Where $\mu$ is mean vector, $C$ is covariance matrix, $|C|$ is the determinant of $C$ matrix, $x \in R^{m}$ is data vector, and $m$ is the dimension of $x$ respectively.

\end{enumerate}

\section{Experiment}
Furthermore, we have provided a set of detailed experiments on the proposed algorithms in different realistic scenarios. However, we maintain the statistical framework provided by the work~\cite{Carter2012} with theoretical guarantees.

We have experimented to determine the usefulness of our algorithm by creating a simulation with 10000000 random vectors with dimensions of 15. The repeated trial shows that our algorithm is not sensitive to initialization seeds and dimensions of the matrix. This requirement was a deciding factor in the choice of the evaluation metric as shown in Equation~\ref{aad_equation}. We have provided more information on the metric in Section~\ref{result-analysis}.

Our experiment will check the effect of varying the size of the initial static window versus the update window as shown in Subsection~\ref{experiment-1} and Subsection~\ref{experiment-2}.

\begin{figure}[H] 
\centering
\includegraphics[scale=0.45]{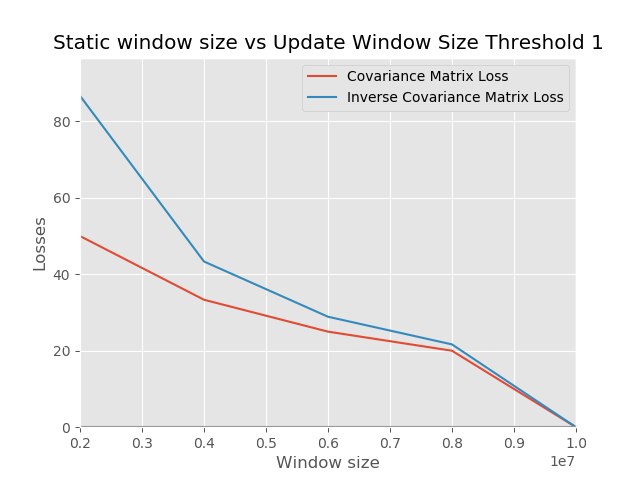}
\caption{Compare the threshold of static vs incremental impact performance of anomaly detection (Version 1)}
\label{experimentchart1}
\end{figure}

\subsection{Experiment 1}
\label{experiment-1}
We evaluate the trade-off between the static window and the update window. The experiment setup is as follows:

\begin{itemize}
\item Split the data into 5 segments train on 1st segment(static), update covariance on 2nd (online), compare with static covariance, and calculate the error.

\item Train on 1, 2 segments (static), update covariance on 3rd (online), compare with static covariance and calculate the error.

\item Train on 1, 2, 3 segments (static), update covariance on 4th (online), compare with static covariance and calculate the error.

\item Train on 1, 2, 3, 4 segments (static), update covariance on 5th (online), compare with static covariance and calculate the error.

\end{itemize}

Figure~\ref{experimentchart1} contains the experimental result. 

\begin{figure}[H] 
\centering
\includegraphics[scale=0.45]{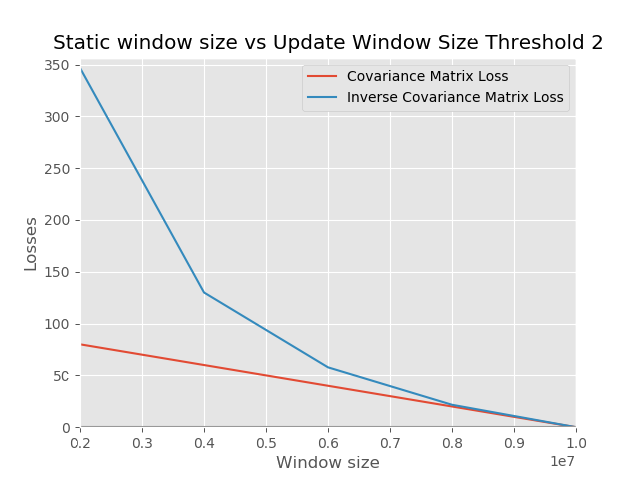}
\caption{Compare the threshold of static vs incremental impact performance of anomaly detection (Version 2)}
\label{experimentchart2}
\end{figure} 

\subsection{Experiment 2}
\label{experiment-2}
The experiment setup is as follows:
\begin{itemize}
\item Split the data into 5 segments.

\item Train on 1st segment(static), update covariance on remaining segments (2,3,4,5) (online), compare with static covariance and calculate error on segments (2,3,4,5)

\item Train on 1, 2 segments (static), update covariance on remaining segments (3,4,5) (online), compare with static covariance and calculate error on segments (3,4,5)

\item Train on 1, 2, 3 segments (static), update covariance on remaining segments (4,5) (online), compare with static covariance and calculate error on segments (4,5)

\item Train on 1, 2, 3, 4 segment(static), update covariance on remaining segments (5) (online), compare with static covariance and calculate error on segments (5)
\end{itemize}

Figure~\ref{experimentchart2} contains the experimental result.

\section{Result Analysis}
\label{result-analysis}
Our random matrices get flattened to a vector and utilized as input. The length of the flattened vector is used as a normalization factor to make the loss metric that is agnostic to the dimension of the matrix. The loss function used in the evaluation is Absolute Average Deviation (AAD) because it gives a tighter bound on the error compared to mean squared error (MSE) or mean absolute deviation (MAD)  as shown in Equation~\ref{aad_equation}. We take the average of the residuals divided by the ground truth for every sample in our evaluation set. If the residual is close to zero, we contribute almost nothing to the measure. On the contrary, if the residual is high, we want to know the difference from the ground truth. 

\begin{equation}
\label{aad_equation}
AAD = \sum_{i=1}^{n} \left| \frac{\hat{Y_i} - Y_i}{Y_i} \right|
\end{equation}

Where $\hat{Y_i}$ is the predicted value, $Y_i$ is the ground truth, and $n$ is the length of the flattened matrix.

We can observe that building your model with more data in the init (static) phase leads to lower errors compared to having fewer data in the init phase and using more of the data for an update. The observation matches our intuition because when you operate in an online mode, you tend to use smaller storage space. However, there is still a performance trade-off when compared to batch mode.

The error at the beginning of our training is significant in both charts. This insight shows that rather than performing the expensive operation of converting a covariance matrix to have the property of positive definiteness, it is better to use random matrices that are positive definite. More data would help us get to convergence as more data arrives.

The success of the experiments has given us the confidence that our multivariate anomaly detection algorithm would have similar characteristics to the univariate case described in the work~\cite{Carter2012}.

\section{Conclusion}
There is no generic anomaly detection that works for every possible task. The underlying assumption in this work is that the features in use capture relevant information about the underlying dynamic of the system. Our proposed implementation is an anomaly detection algorithm for handling multivariate streams even with challenging shifts. In future work, we will make extensions to support non-stationary distributions in multivariate data streams.

\bibliographystyle{ACM-Reference-Format}
\bibliography{sample-sigconf}

\end{document}